\documentclass[runningheads]{llncs}

\usepackage{eccv}

\usepackage{comment}
\usepackage{multirow}
\usepackage{eccvabbrv}

\usepackage{graphicx}
\usepackage{booktabs}

\usepackage[accsupp]{axessibility}  

\usepackage[pagebackref,breaklinks,colorlinks,citecolor=eccvblue]{hyperref}

\usepackage{orcidlink}

\begin{document}

\title{Unsupervised Intrinsic Image Decomposition with LiDAR Intensity Enhanced Training} 

\titlerunning{LIET for IID}

\author{
Shogo Sato \inst{1}\orcidlink{0000-0003-3070-5880} \and 
Takuhiro Kaneko \and
Kazuhiko Murasaki \and
Taiga Yoshida \and
Ryuichi Tanida\and
Akisato Kimura
}

\authorrunning{S.~Sato et al.}

\institute{NTT Corporation, 1-1, Hikarino-oka, Yokosuka-shi, Kanagawa, Japan}

\maketitle

\begin{abstract}
Unsupervised intrinsic image decomposition (IID) is the process of separating a natural image into albedo and shade without these ground truths. A recent model employing light detection and ranging (LiDAR) intensity demonstrated impressive performance, though the necessity of LiDAR intensity during inference restricts its practicality. Thus, IID models employing only a single image during inference while keeping as high IID quality as the one with an image plus LiDAR intensity are highly desired. To address this challenge, we propose a novel approach that utilizes only an image during inference while utilizing an image and LiDAR intensity during training. Specifically, we introduce a partially-shared model that accepts an image and LiDAR intensity individually using a different specific encoder but processes them together in specific components to learn shared representations. In addition, to enhance IID quality, we propose albedo-alignment loss and image-LiDAR conversion (ILC) paths. Albedo-alignment loss aligns the gray-scale albedo from an image to that inferred from LiDAR intensity, thereby reducing cast shadows in albedo from an image due to the absence of cast shadows in LiDAR intensity. Furthermore, to translate the input image into albedo and shade style while keeping the image contents, the input image is separated into style code and content code by encoders. The ILC path mutually translates the image and LiDAR intensity, which share content but differ in style, contributing to the distinct differentiation of style from content. Consequently, LIET achieves comparable IID quality to the existing model with LiDAR intensity, while utilizing only an image without LiDAR intensity during inference.
\end{abstract}

\section{Introduction}
Intrinsic image decomposition (IID) is the process of separating a natural image into albedo and shade within Lambertian scenes. IID provides benefits for various high-level computer vision tasks, such as texture editing~\cite{beigpour2011object,meka2016live} and semantic segmentation~\cite{upcroft2014lighting,wang2017robust,baslamisli2018joint}. The origins of IID can be traced back to early work in computer vision during the 1970s~\cite{land1971,barrow1978}, where researchers grappled with the challenge of recovering albedos from shaded images. Since IID is an ill-posed problem that separates a natural image into its albedo and shade, researchers 

\newpage
\begin{figure}
\centering
\includegraphics[width=1.0\linewidth]{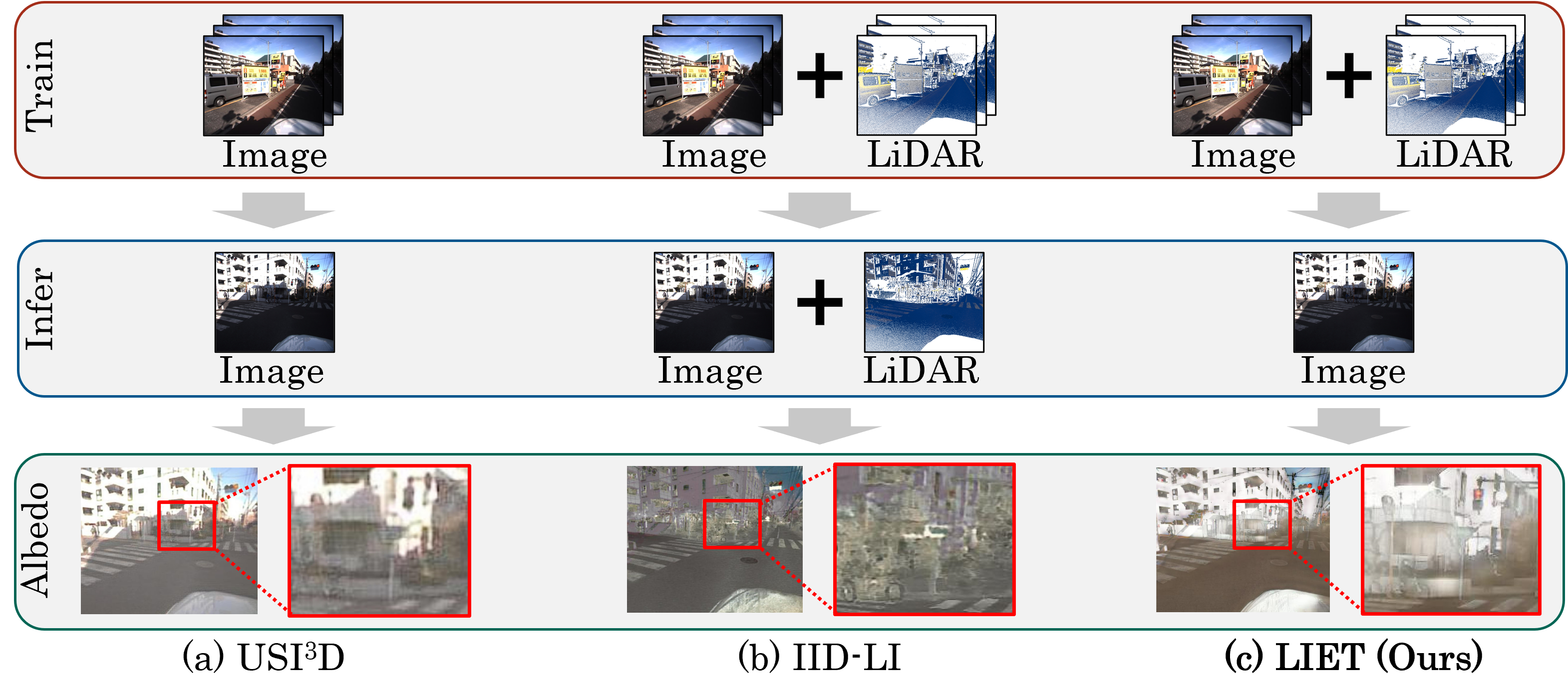}
\caption{\label{fig1} Train/infer schemes and examples of inferred albedos by (a) USI$^3$D~\cite{liu2020}, (b) IID-LI~\cite{Sato2023}, and (c) LIET (our proposed model). USI$^3$D, utilizing only a single image for both training and inference, leaves cast shadows on the inferred albedo. On the other hand, IID-LI utilizes LiDAR intensity during training and inference, making shadows further less noticeable. However, IID-LI has restricted applicability due to the requirement of LiDAR intensity even during inference. LIET utilizes only an image for inference to expand its usage scenarios, and utilizes both an image and its corresponding LiDAR intensity during training to make shadows less noticeable.}
\end{figure}
\vspace{-0.5cm}

\noindent
have traditionally addressed this challenge by incorporating a variety of priors, including albedo flatness, shade smoothness~\cite{land1971,weiss2001,grosse2009,zhao2012,bell2014,bi2015} and dependence between shade and geometry~\cite{chen2013, lee2012, jeon2014}. Recently, a notable development has been the emergence of supervised learning models~\cite{narihira2015_1,zhou2015,narihira2015_2,krahenbuhl2018,fan2018,zhou2019glosh,luo2020,zhu2021}, trained on the ground-truth albedo and shade corresponding to an input image with sparsely-annotated datasets~\cite{bell2014} or synthetic datasets~\cite{butler2012,chang2015,li2018cg}. However, supervised learning models are not ideal since it is difficult to prepare ground truths by eliminating illumination from images within general scenes. On the other hand, unsupervised learning models~\cite{ma2018,li2018,liu2020,seo2021}, that do not utilize ground-truth albedo and shade corresponding to the input image, encountered IID-quality limitations as shown in~\cref{fig1} (a), particularly in their capacity to reduce cast shadows. More recently, an unsupervised learning model that utilizes light detection and ranging (LiDAR) intensity, called intrinsic image decomposition with LiDAR intensity (IID-LI)~\cite{Sato2023}, has notably enhanced IID quality as shown in~\cref{fig1} (b). LiDAR intensity refers to the strength of light reflected from object surfaces, and is equivalent to albedo in infrared wavelength. Thus, LiDAR intensity is effective for IID tasks. However, the applicability of IID-LI is limited since it requires LiDAR intensity even during inference.

This paper aims to employ only a single image without LiDAR intensity during inference to expand usage scenarios while keeping the high IID quality demonstrated by IID-LI as shown in~\cref{fig1} (c). To accomplish this objective, we propose \textit{unsupervised single-image intrinsic image decomposition with LiDAR intensity enhanced training (LIET)}. 
In the previous IID-LI framework, completely-shared model accepts both an image and LiDAR intensity as input and processes them simultaneously for both the training and inference processes. On the other hand, our proposed LIET is implemented with a \textit{partially-shared model} that accepts an image and LiDAR intensity individually using a different specific encoder but processes them together in specific components to learn shared representations. The inference from a single image is achieved by utilizing only image-encoder path during inference, while both of the image-encoder path and LiDAR-encoder path are utilized during training. Furthermore, to enhance the IID quality, we propose \textit{albedo-alignment loss} and \textit{image-LiDAR conversion (ILC) paths}. The albedo-alignment loss aligns the gray-scaled albedo from an image to that inferred from its corresponding LiDAR intensity. LiDAR intensity reflects the object-surface properties independent from sunlight conditions and cast shadows, hence the albedo from LiDAR intensity has the potential to be a criterion for reducing cast shadow. In addition, to translate the input image into albedo and shade style while keeping the image contents \footnote{Style and contents represent domain-variant components such as illuminations, and domain-invariant components such as object edges, respectively}, the input image is separated into style and content codes by encoders. Due to the shared content but differing styles between an image and its corresponding LiDAR intensity, the ILC paths that mutually translate them facilitate separating into content and style codes, enhancing the IID quality. 

The performance of LIET is investigated by comparing it with existing IID models including energy optimization models~\cite{grosse2009,bell2014,bi2015}, weakly supervised model~\cite{fan2018}, and unsupervised models~\cite{li2018, Lettry2018,liu2020, Sato2023} in IID quality metrics and image quality assessment (IQA)~\cite{NIQA, TReS, MUSI, HIQA, DBCN} on inferred albedos. The main contributions of this study are summarized as follows.
\begin{itemize}
  \item We propose \textit{unsupervised single-image intrinsic image decomposition with LiDAR intensity enhanced training (LIET)} with a \textit{partially-shared model} that accepts an image and LiDAR intensity individually using a different specific encoder but processes them together in specific components to learn shared representations. This model architecture reaches a single image inference, while both an image and LiDAR intensity are utilized during training.
  \item To enhance the effective utilization of LiDAR intensity, we introduce \textit{albedo-alignment loss} to align the albedo inferred from an image to that from its corresponding LiDAR intensity, and \textit{image-LiDAR conversion (ILC) paths} to translate the input image into albedo and shade style while keeping the image contents. Additionally, the ablation study demonstrates the effectiveness of each proposed architecture and loss.
  \item In terms of IID quality, our proposed model, which employs only an image during inference demonstrates comparable performance to the existing model, which employs both an image and LiDAR intensity during inference. 
\end{itemize}

\section{Related work}\label{sec2}
This section initially introduces general image-to-image translation (I2I) models that translate an input image from their source domain to the target domain. Since IID represents a specific form of I2I that translates an input image from the image domain into albedo and shade domains, several IID models and LIET are based on the I2I framework. Furthermore, we describe the existing unsupervised IID models and examples of LiDAR intensity utilization in this section.\smallskip\\
\textbf{Image-to-image translation.}
I2I models are designed to translate an input image from their source domain to the target domain. Most of the I2I models rely on deep generative models, and typical models include generative adversarial networks (GAN)~\cite{goodfellow2014generative} such as pix2pix~\cite{isola2017image}. Due to the challenge of acquiring paired images for each domain, CycleGAN~\cite{zhu2017unpaired} was proposed as an I2I model that does not require paired images. Additionally, unsupervised image-to-image translation networks (UNIT)~\cite{liu2017unsupervised} achieved unsupervised I2I by implementing weight sharing within the latent space. Conversely, UNIT and CycleGAN require training as many models as the number of domains to be translated, leading to high computational costs. Thus, StarGAN~\cite{choi2018stargan} and multimodal unsupervised image-to-image translation (MUNIT)~\cite{huang2018multimodal} were introduced to translate images into multiple domains using a single model. Additionally, diverse image-to-image translation via disentangled representations (DRIT)~\cite{lee2018diverse}, which amalgamates the advantages of UNIT and MUNIT, was presented. More recently, diffusion model~\cite{ho2020denoising,saharia2022image} began to be applied for I2I~\cite{li2023bbdm,cheng2023general}. USI$^3$D~\cite{liu2020}, IID-LI~\cite{Sato2023}, and LIET are implemented based on MUNIT as specialized models for the IID task.\smallskip\\
\textbf{Unsupervised learning models for IID.}
Acquiring ground-truth albedo and shade corresponding to an input image in general scenes presents a considerable challenge, thus necessitating the use of unsupervised learning models that do not depend on such ground truths. To facilitate IID in the absence of ground truths, two primary strategies are employed, one using multiple images captured under different conditions and the other using a single image and synthetic data for albedo and shade domains that do not correspond to the image. As the first strategy, models have been trained using pairs of images under varying illumination conditions~\cite{ma2018}, as well as sequences of related images~\cite{yu2019}. More recent models~\cite{srinivasan2021nerv,boss2021nerd,zhang2021physg,zhang2021nerfactor,boss2021neural,munkberg2022extracting,hasselgren2022shape,zhu2023i2} leveraging the neural radiance field (NeRF)~\cite{mildenhall2021nerf} framework from multi-view images have been proposed. Conversely, as the second strategy, USI$^3$D~\cite{liu2020} employs an image for IID and the synthetic albedo and shade domain data for ensuring albedo or shade domain likelihood, resulting in the enhancement of the decomposition quality without direct ground truth. In addition, IID-LI~\cite{Sato2023} has incorporated LiDAR intensity based on USI$^3$D to reduce cast shadows and demonstrated impressive IID performance. However, IID-LI has restricted applicability due to the requirement for LiDAR intensity even during inference. Thus, IID models employing only a single image during inference while keeping as high IID quality as IID-LI are highly desired.\smallskip\\
\textbf{LiDAR intensity utilization.}
LiDAR is a device for measuring the distance to the object surfaces based on the time of flight from infrared laser irradiation to the reception of reflected light. In addition to the distance measurement, LiDAR also captures the intensity of the reflected light from the object surfaces, commonly referred to as LiDAR intensity. This LiDAR intensity is unaffected by variations in sunlight conditions or shading while preserving the texture of 
\newpage

\begin{figure}
\centering
\includegraphics[width=1.0\linewidth]{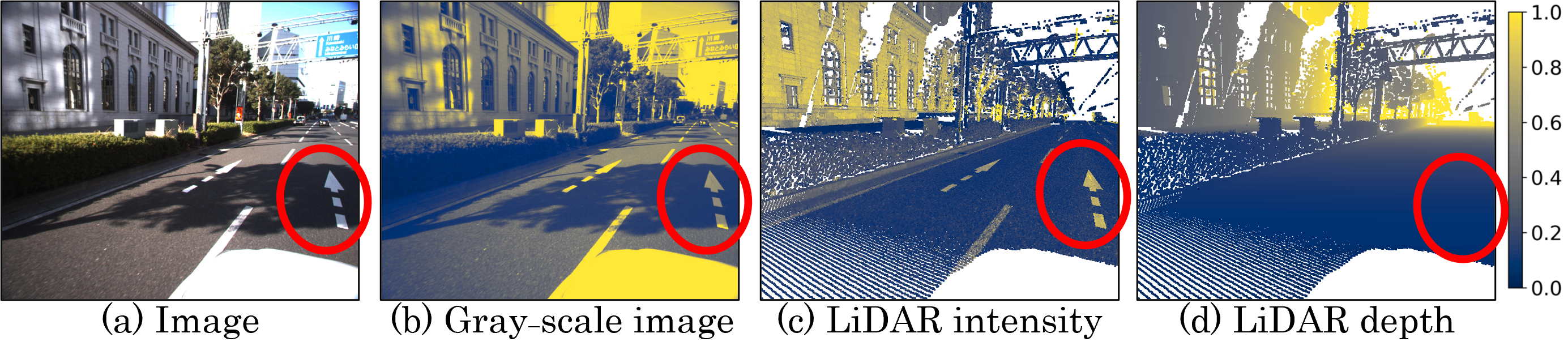}
\caption{\label{fig2}Examples of (a) input image, (b) gray-scale image, and its corresponding (c) LiDAR intensity and (d) LiDAR depth. The red circle indicates the regions with cast shadows and white arrows. The shadow and the white arrow are visible in gray-scale image. LiDAR intensity has no cast shadows while maintaining white arrows, since LiDAR intensity is calculated from the intensity ratio of irradiated and reflected lights, equivalent to an albedo at infrared wavelength. Conversely, LiDAR depth represents the distance to objects, resulting in the absence of cast shadows and white arrows.}
\end{figure}
\vspace{-0.5cm}

\noindent
object surfaces as illustrated in~\cref{fig2}. Thus, LiDAR intensity has the potential for effective utilization in the context of IID tasks. Also, LiDAR intensity is widely applied due to its ability to depict surface properties, for instance, shadow detection~\cite{guislain,sato2023shadow}, hyper-spectral data correction~\cite{priem2016,brell2017}, and object recognition~\cite{lang2009,kashani2015,man2015,li2023memoryseg}. Note that LiDAR and a camera typically operate in distinct wavelength bands; the near-infrared band and the visible light band, respectively. Hence, LiDAR intensity is not to be used directly as is for ground-truth albedo. LIET approaches the difference in wavelength by calculating loss with albedo inferred from LiDAR intensity, rather than directly utilization.

\section{Proposed model (LIET)}\label{sec4}
\subsection{Problem formulation}
This section describes the problem formulation addressed by LIET framework. Initially, the inference process employs only a real-world image $x_{\rm{I}}$ to infer albedo $x_{\rm{RI}}$ and shade $x_{\rm{SI}}$. Meanwhile in the training process, a set of a real-world image $x_{\rm{I}}$ and its corresponding LiDAR intensity $x_{\rm{L}}$ is prepared. The ground-truth albedo and shade corresponding to the image are not prepared due the difficulty of obtaining ground truths in the real world. Instead of these ground truths, albedo $x_{\rm{R}}$ and shade $x_{\rm{S}}$ derived from synthetic dataset that do not corresponding to the image $x_{\rm{I}}$ are prepared, thereby enabling calculation of the distributions for albedo and shade.

\subsection{USI$^3$D architecture}
\begin{figure}
\centering
\includegraphics[width=0.9\linewidth]{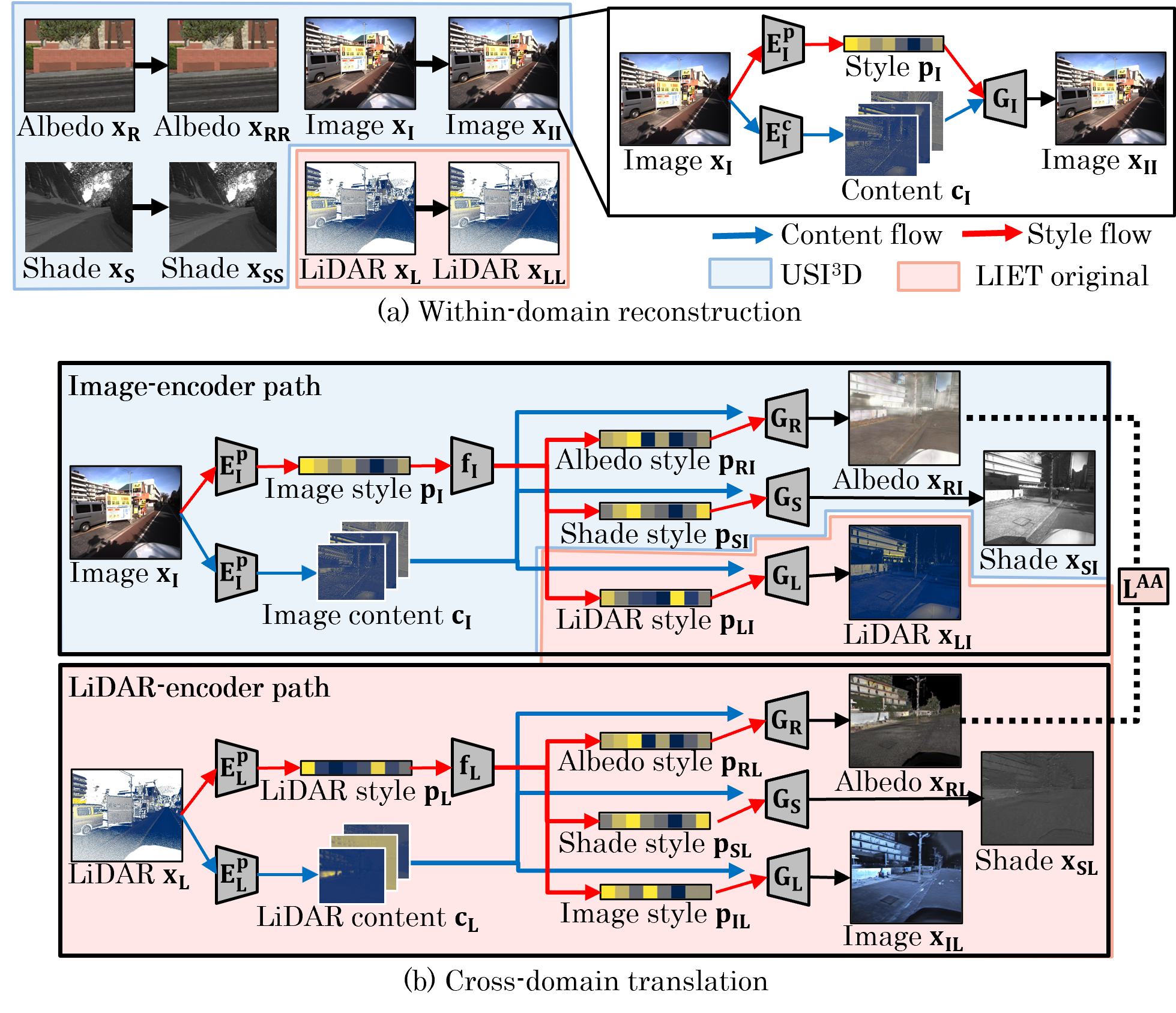}
\caption{\label{fig3} LIET architecture including (a) within-domain reconstruction and (b) cross-domain translation. (a) For each domain (image ${\rm{I}}$, LiDAR intensity ${\rm{L}}$, albedo ${\rm{R}}$, shade ${\rm{S}}$), an input $x_{\rm{X}}$ is fed into style $E^p_{\rm{X}}$ and content $E^c_{\rm{X}}$ encoders to calculate style $p_{\rm{X}}$ and content $c_{\rm{X}}$ codes for domain $X\in\{\rm{I, L, R, S}\}$. These codes are used at generators $G_{\rm{X}}$ to reconstruct the inputs within their domains. (b) The image-encoder path accepts an image $x_{\rm{I}}$ and infers image style $p_{\rm{I}}$ and content $c_{\rm{I}}$ codes as within-domain reconstruction. Subsequently, $p_{\rm{I}}$ is fed into a style mapping function $f_{\rm{I}}$ to yield domain-specific style codes ($p_{\rm{RI}}$, $p_{\rm{SI}}$, $p_{\rm{LI}}$) for generating respective domains via generators ($G_{\rm{R}}, G_{\rm{S}}, G_{\rm{L}}$). Similarly, the LiDAR-encoder path uses $x_{\rm{L}}$ to infer albedo $x_{\rm{RL}}$, shade $x_{\rm{SL}}$, and image $x_{\rm{IL}}$ through LiDAR style and content encoders ($E^p_{\rm{L}}, E^c_{\rm{L}}$). The albedo-alignment loss $\mathcal{L}^{\rm{AA}}$ aligns the gray-scaled albedo from an image to that inferred from the LiDAR intensity for reducing cast shadows. 
}
\end{figure}

\textbf{Overview.}
USI$^3$D consists of within-domain reconstruction and cross-domain translation as illustrated in light-blue regions of~\cref{fig3}. The within-domain reconstruction aims to extract features for each domain of image $x_{\rm{I}}$, albedo $x_{\rm{R}}$, and shade $x_{\rm{S}}$ by encoders and decoders. The cross-domain translation infers albedo $x_{\rm{RI}}$ and shade $x_{\rm{SI}}$ from an input image $x_{\rm{I}}$.\smallskip\\
\textbf{Within-domain reconstruction.}
As illustrated in~\cref{fig3} (a), for each domain 
\newpage

\vspace{-0.5cm}
\noindent
including image ${\rm{I}}$, albedo ${\rm{R}}$, and shade ${\rm{S}}$, an input $x_{\rm{X}}$ is fed into style encoder $E^p_{\rm{X}}$ and content encoder $E^c_{\rm{X}}$ to derive style code $p_{\rm{X}}$ and content code $c_{\rm{X}}$, respectively, for $X\in \{\rm{I}, \rm{R}, \rm{S}\}$. These codes are input to domain-specific generators $G_{\rm{X}}$, reconstructing the original inputs within their respective domains $x_{\rm{XX}}$. For example, for an image $x_{\rm{I}}$, the image-style encoder $E^p_{\rm{I}}$ and image-content encoder $E^c_{\rm{I}}$ are utilized to extract image-style code $p_{\rm{I}}$ and image-content code $c_{\rm{I}}$, respectively, with the reconstruction achieved by image generator $G_{\rm{I}}(c_{\rm{I}},p_{\rm{I}})$. Analogous processes apply for albedo, and shade reconstructions.\smallskip\\
\textbf{Cross-domain translation.}
For cross-domain translation, an input image $x_{\rm{I}}$ is processed to infer albedo $x_{\rm{RI}}$ and shade $x_{\rm{SI}}$. First, an input $x_{\rm{I}}$ is fed into image-style encoder $E^p_{\rm{I}}$ and image-content encoder $E^c_{\rm{I}}$ to derive image-style code $p_{\rm{I}}$ and image-content code $c_{\rm{I}}$, respectively. A style mapping function $f_{\rm{I}}$ then adjusts $p_{\rm{I}}$ to generate domain-specific style codes for albedo $p_{\rm{RI}}$ and shade $p_{\rm{SI}}$. Subsequently, these style codes, alongside $c_{\rm{I}}$, are input into their respective generators ($G_{\rm{R}}(c_{\rm{I}},p_{\rm{RI}}), G_{\rm{S}}(c_{\rm{I}},p_{\rm{SI}})$) to infer the albedo $x_{\rm{RI}}$ and shade $x_{\rm{SI}}$, respectively. Note that all encoders and decoders are shared between within-domain reconstruction and cross-domain translation. \smallskip\\
\textbf{Adversarial training.} 
To improve the translation quality from the source domain into the target domain, the translated images are input into discriminators. Discriminators for albedo $D_{\rm{R}}$ and shade $D_{\rm{S}}$ are implemented for adversarial training. For instance, the albedo inferred from image $x_{\rm{RI}}$ and the albedo domain data $x_{\rm{R}}$ are provided into the albedo discriminator $D_{\rm{R}}$ to evaluate the domain likelihood.

\subsection{LIET architectures}
\textbf{Overview.}
LIET is based on USI$^3$D and consists of within-domain reconstruction and cross-domain translation as illustrated in \cref{fig3}. For within-domain reconstruction, LiDAR intensity $x_{\rm{L}}$ reconstruction is added to image $x_{\rm{I}}$, albedo $x_{\rm{R}}$, and shade $x_{\rm{S}}$ reconstructions. For cross-domain translation, LIET introduces a LiDAR-encoder path by employing a specific encoders ($E^p_{\rm{L}}, E^c_{\rm{L}}$) to infer albedo $x_{\rm{RL}}$ and shade $x_{\rm{SL}}$ from LiDAR intensity $x_{\rm{L}}$, while USI3D had only an image-encoder path. The inference from an image is achieved by utilizing only the image-encoder path during inference, while both paths are utilized during training. To enhance the IID quality, we introduce a albedo-alignment loss to align the albedo inferred from an image $x_{\rm{RI}}$ to that from its corresponding LiDAR intensity $x_{\rm{RL}}$. Furthermore, image-LiDAR conversion (ILC) paths that mutually translate image and LiDAR intensity are also implemented for effective utilization of LiDAR intensity. \smallskip\\
\textbf{Within-domain reconstruction.}
In the same manner as USI$^3$D, within-domain reconstructions for image, albedo, and shade are implemented. In addition, reconstruction for LiDAR intensity $x_{\rm{L}}$ is also implemented in LIET as shown in~\cref{fig3} (a), to extract feature of LiDAR intensity. The LiDAR intensity $x_{\rm{L}}$ is input into LiDAR-style encoder $E^p_{\rm{L}}$ and LiDAR-content encoder $E^c_{\rm{L}}$ to calculate the LiDAR-style code $p_{\rm{L}}$ and LiDAR-content code $c_{\rm{L}}$, respectively. Subsequently, LiDAR generator $G_{\rm{L}}(c_{\rm{L}},p_{\rm{L}})$ is utilized to reconstruct the LiDAR intensity $x_{\rm{LL}}$.\smallskip\\
\textbf{Cross-domain translation.}
As illustrated in the light-blue region of~\cref{fig3} (b), USI$^3$D infers albedo $x_{\rm{RI}}$ and shade $x_{\rm{SI}}$ from an image $x_{\rm{I}}$. Within the LIET framework, the objective is to maintain the use of solely an image $x_{\rm{I}}$ as inputs during inference, while simultaneously leveraging both the image $x_{\rm{I}}$ and LiDAR intensity $x_{\rm{L}}$ during training. To this end, alongside the conventional image-encoder path, a LiDAR-encoder path employs LiDAR intensity $x_{\rm{L}}$ to infer albedo $x_{\rm{RL}}$ and shade $x_{\rm{SL}}$. The LiDAR intensity $x_{\rm{L}}$ is fed into LiDAR-style encoder $E^p_{\rm{L}}$ and LiDAR-content encoder $E^c_{\rm{L}}$ to calculate the LiDAR-style $p_{\rm{L}}$ and LiDAR-content $c_{\rm{L}}$ codes, respectively. The LiDAR-style $p_{\rm{L}}$ is input into the style mapping function $f_{\rm{L}}$ to infer style codes for albedo $p_{\rm{RL}}$ and shade $p_{\rm{SL}}$. Finally, these style and content codes are fed into generators ($G_{\rm{R}}(c_{\rm{L}},p_{\rm{RL}}), G_{\rm{S}}(c_{\rm{L}},p_{\rm{SL}})$) to infer albedo $x_{\rm{RL}}$ and shade $x_{\rm{SL}}$. This partially-shared model supports the concurrent objectives of effectively leveraging LiDAR intensity during training and maintaining exclusive reliance on an image input during inference. \smallskip\\
\textbf{Image-LiDAR conversion (ILC) paths.}
As illustrated in the light-blue region of~\cref{fig3} (b), USI$^3$D separates an image $x_{\rm{I}}$ into content code $c_{\rm{I}}$ and style code $p_{\rm{I}}$ to infer albedo $x_{\rm{RI}}$ and shade $x_{\rm{SI}}$, and this separation process is critical as it directly impacts the IID quality. Since USI$^3$D utilizes the independence datasets for image $x_{\rm{I}}$, albedo $x_{\rm{R}}$ and shade $x_{\rm{S}}$ without shared content, this separation relies solely on the style information unique to each domain. On the other hand, in our problem formulation, LiDAR intensity $x_{\rm{L}}$ corresponding to the image $x_{\rm{I}}$ is also available. This helps us separate an image into content and style codes $(c_{\rm{I}}, p_{\rm{I}})$ accurately since the image and the corresponding LiDAR intensity should share the content. LIET incorporates the decoder $G_{\rm{L}}(c_{\rm{I}}, p_{\rm{LI}})$ for inferring the LiDAR intensity in the image-encoder path, thereby, the content code $c_{\rm{I}}$ and style code $p_{\rm{I}}$ of the input image are effortlessly separated. Furthermore, to enhance the inference quality of albedo $x_{\rm{RL}}$ and shade $x_{\rm{SL}}$ from the LiDAR intensity $x_{\rm{L}}$, the image inference path is also added to the LiDAR-encoder path in the same manner as the image to LiDAR intensity path. \smallskip\\
\textbf{Adversarial training.} 
Similar to USI$^3$D, inferred albedo $x_{\rm{RI}}$ and shade $x_{\rm{SI}}$ from an image $x_{\rm{I}}$ is input into albedo discriminator $D_{\rm{R}}$ and shade discriminator $D_{\rm{S}}$, respectively, to improve the translation quality from the source domain into the target domain. In LIET, due to the incorporation of the LiDAR-encoder path, the albedo $x_{\rm{RL}}$ and shade $x_{\rm{SL}}$ inferred from LiDAR intensity $x_{\rm{L}}$ are also input into their respective discriminators ($D_{\rm{R}}, D_{\rm{S}}$). Furthermore, along with the ILC paths, inferred LiDAR intensity $x_{\rm{LI}}$ and image $x_{\rm{IL}}$ are also evaluated by LiDAR-intensity discriminator $D_{\rm{L}}$ and image discriminator $D_{\rm{I}}$, respectively. Note that all encoders and decoders are shared in within-domain reconstruction, the image-encoder path, and the LiDAR-encoder path.

\subsection{Losses}
\begin{figure}
\centering
\includegraphics[width=0.6\linewidth]{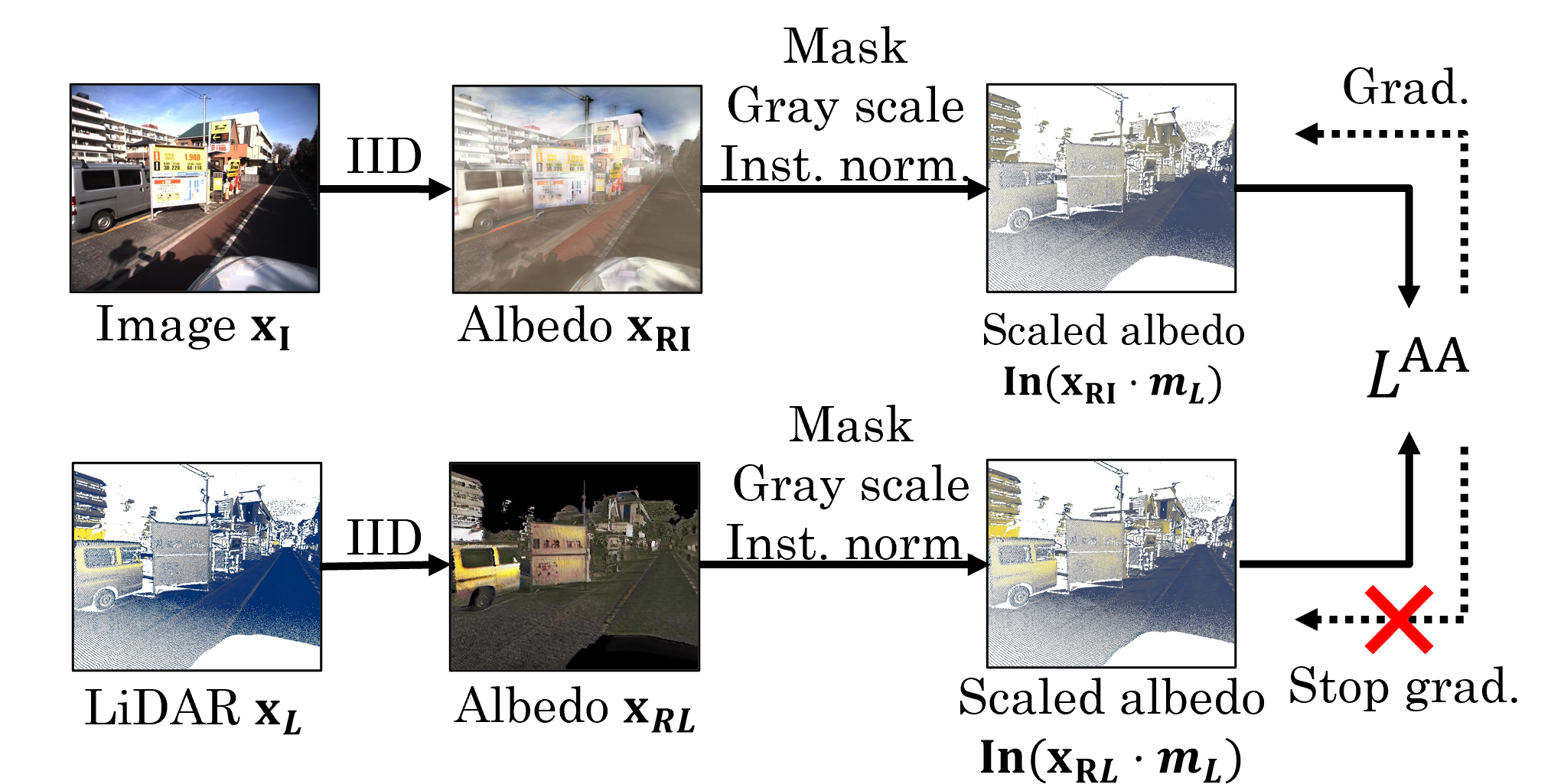}
\caption{\label{fig5} Calculation process of albedo-alignment loss $\mathcal{L}^{\rm{AA}}$. First, albedo from an image $x_{\rm{RI}}$ and that from its corresponding LiDAR intensity $x_{\rm{RL}}$ are computed. Subsequently, these albedos are masked to the points with LiDAR values and then gray scale. Next, instance normalization is performed to align the scales of these albedos, and the distance between these scaled albedos is calculated. A stop gradient is performed on the LiDAR-encoder path side to align $x_{\rm{RI}}$ to $x_{\rm{RL}}$ since the LiDAR intensity is independent of sunlight conditions. LiDAR intensity, scaled albedo from the image, and that from LiDAR intensity are represented in a cividis color map due to their gray scale.}
\end{figure}

In this section, we describe the loss functions computed during within-domain reconstruction and cross-domain translation.\smallskip\\
\textbf{Image reconstruction loss $\mathcal{L}^{\rm{img}}$.}
Initially, the input images should be reconstructed after passing through the within-domain reconstruction process, hence image reconstruction loss $\mathcal{L}^{\rm{img}}$ is defined in~\cref{eq1}.
\begin{equation}\label{eq1}
  \mathcal{L}^{\rm{img}}=
  \|x_{\rm{II}}-x_{\rm{I}}\|+
  \|x_{\rm{LL}}-x_{\rm{L}}\|+
  \|x_{\rm{RR}}-x_{\rm{R}}\|+
  \|x_{\rm{SS}}-x_{\rm{S}}\|,
\end{equation}
where $x_{\rm{II}}$, $x_{\rm{LL}}$, $x_{\rm{RR}}$, and $x_{\rm{SS}}$ are reconstructed images by within-domain reconstruction for image, LiDAR intensity, albedo, and shade, respectively.\smallskip\\
\textbf{Style reconstruction loss $\mathcal{L}^{\rm{sty}}$ and content code reconstruction loss $\mathcal{L}^{\rm{cnt}}$.} Since the reconstructed images should maintain their styles and contents, the style reconstruction loss $\mathcal{L}^{\rm{sty}}$ and content code reconstruction loss $\mathcal{L}^{\rm{cnt}}$ are defined in~\cref{eq2} and~\cref{eq3}, respectively.
\begin{multline}\label{eq2}
  \mathcal{L}^{\rm{sty}}=
  \|E^p_{\rm{L}}(x_{\rm{LI}})-p_{\rm{LI}}\|+
  \|E^p_{\rm{R}}(x_{\rm{RI}})-p_{\rm{RI}}\|+
  \|E^p_{\rm{S}}(x_{\rm{SI}})-p_{\rm{SI}}\|\\
 +\|E^p_{\rm{I}}(x_{\rm{IL}})-p_{\rm{IL}}\|+
  \|E^p_{\rm{R}}(x_{\rm{RL}})-p_{\rm{RL}}\|+
  \|E^p_{\rm{S}}(x_{\rm{SL}})-p_{\rm{SL}}\|,
\end{multline}
\begin{multline}\label{eq3}
  \mathcal{L}^{\rm{cnt}}=
  \|E^c_{\rm{L}}(x_{\rm{LI}})-c_{\rm{I}}\|+
  \|E^c_{\rm{R}}(x_{\rm{RI}})-c_{\rm{I}}\|+
  \|E^c_{\rm{S}}(x_{\rm{SI}})-c_{\rm{I}}\|\\
 +\|E^c_{\rm{I}}(x_{\rm{IL}})-c_{\rm{L}}\|+
  \|E^c_{\rm{R}}(x_{\rm{RL}})-c_{\rm{L}}\|+
  \|E^c_{\rm{S}}(x_{\rm{SL}})-c_{\rm{L}}\|.
\end{multline}
\textbf{Adversarial loss $\mathcal{L}^{\rm{adv}}$.} 
Moreover, the adversarial loss $\mathcal{L}^{\rm{adv}}$~\cite{goodfellow2014generative} is defined as~\cref{eq4} to ensure that the image inferred through cross-domain translation aligns with the distribution of the target domain.
\begin{multline}\label{eq4}
  \mathcal{L}^{\rm{adv}}=
   \log(1-D_{\rm{R}}(x_{\rm{RI}}))+\log(D_{\rm{R}}(x_{\rm{R}}))
  +\log(1-D_{\rm{R}}(x_{\rm{RL}}))+\log(D_{\rm{R}}(x_{\rm{R}}))\\
  +\log(1-D_{\rm{S}}(x_{\rm{SI}}))+\log(D_{\rm{S}}(x_{\rm{S}}))
  +\log(1-D_{\rm{S}}(x_{\rm{SL}}))+\log(D_{\rm{S}}(x_{\rm{S}}))\\
  +\log(1-D_{\rm{L}}(x_{\rm{LI}}))+\log(D_{\rm{L}}(x_{\rm{L}}))
  +\log(1-D_{\rm{I}}(x_{\rm{IL}}))+\log(D_{\rm{I}}(x_{\rm{I}})).
\end{multline}
\textbf{VGG loss $\mathcal{L}^{\rm{VGG}}$.} 
To preserve the object edges and colors of the input image, the distance between the input image and the inferred albedo within the VGG feature space is computed~\cite{karen2015very,chen2017photographic,wang2018high} for the VGG loss $\mathcal{L}^{\rm{VGG}}$~\cite{johnson2016perceptual} in \cref{eq5}.
\begin{equation}\label{eq5}
  \mathcal{L}^{\rm{VGG}}=\|V(x_{\rm{I}})-V(x_{\rm{RI}})\|,
\end{equation}
where $V$ is pre-trained visual-perception network such as VGG-19~\cite{karen2015very}. \smallskip\\
\textbf{KLD loss $\mathcal{L}^{\rm{KLD}}$.}
Additionally, the Kullback-Leibler divergence (KLD) loss $\mathcal{L}^{\rm{KLD}}$ is formulated as \cref{eq6} to align the probability distributions of inferred albedo style $q(p_{\rm{RI}})$ and shade style $q(p_{\rm{SI}})$ from an image with those calculated from synthetic data $(q(p_{\rm{R}}), q(p_{\rm{S}}))$ facilitated by a style mapping function.
\begin{equation}\label{eq6}
  \mathcal{L}^{\rm{KLD}}=\|\log q(p_{\rm{RI}})-\log q(p_{\rm{R}})\|+\|\log q(p_{\rm{SI}})-\log q(p_{\rm{S}})\|.
\end{equation}
\textbf{Physical loss $\mathcal{L}^{\rm{phy}}$.}
Given the assumption of a Lambertian surface in the IID task, the product of albedo and shade is expected to match the input image. Thus physical loss $\mathcal{L}^{\rm{phy}}$ is defined in \cref{eq7}.
\begin{equation}\label{eq7}
  \mathcal{L}^{\rm{phy}}=\|x_{\rm{I}}-x_{\rm{RI}}\cdot x_{\rm{SI}}\|.
\end{equation}
\textbf{Albedo-alignment loss $\mathcal{L}^{\rm{AA}}$.}
To improve the IID quality, we propose albedo-alignment loss $\mathcal{L}^{\rm{AA}}$ as depicted in~\cref{fig5},
aligning the gray-scale albedo from an image to that inferred from its corresponding LiDAR intensity. Since the albedo inferred from LiDAR intensity $x_{\rm{RL}}$ is independent of daylight conditions and cast shadows, the IID quality is expected to improve by aligning the albedo inferred from an image $x_{\rm{RI}}$ to that inferred from its corresponding LiDAR intensity $x_{\rm{RL}}$. Additionally, these albedos are required to compare in gray scale due to the lack 

\newpage

\vspace{-0.5cm}
\noindent
of hue in LiDAR intensity. Thus, albedo-alignment loss $\mathcal{L}^{\rm{AA}}$ is defined in~\cref{eq8} to compute the distance between $x_{\rm{RI}}$ and $x_{\rm{RL}}$ in gray scale.
\begin{equation}\label{eq8}
  \mathcal{L}^{\rm{AA}}=\|\rm{In}(x_{\rm{RI}}\cdot m_{\rm{L}})-\rm{In}(x_{\rm{RL}}\cdot m_{\rm{L}})\|,
\end{equation}
where $m_{\rm{L}}$ represents the mask denoting the presence of LiDAR intensity values. $\rm{In}(\cdot)$ is an instance normalization~\cite{ulyanov2016instance} and gray-scale function. The instance normalization is used to align the scales of $x_{\rm{RI}}$ and $x_{\rm{RL}}$. In addition, a stop gradient is performed on the LiDAR-encoder path side to align $x_{\rm{RI}}$ to $x_{\rm{RL}}$.

In summary, LIET optimizes the loss function in \cref{eq9}.
\begin{multline}\label{eq9}
  \mathcal{L}^{\rm{LIET}}= \mathcal{L}^{\rm{adv}}
  +\lambda_{\rm{img}}\mathcal{L}^{\rm{img}}
  +\lambda_{\rm{sty}}\mathcal{L}^{\rm{sty}}
  +\lambda_{\rm{cnt}}\mathcal{L}^{\rm{cnt}}\\
  +\lambda_{\rm{KLD}}\mathcal{L}^{\rm{KLD}}
  +\lambda_{\rm{VGG}}\mathcal{L}^{\rm{VGG}}
  +\lambda_{\rm{phy}}\mathcal{L}^{\rm{phy}}
  +\lambda_{\rm{AA}}\mathcal{L}^{\rm{AA}}.
\end{multline}
$\lambda_{\rm{img}}$, $\lambda_{\rm{sty}}$, $\lambda_{\rm{cnt}}$, $\lambda_{\rm{KLD}}$, $\lambda_{\rm{VGG}}$, $\lambda_{\rm{phy}}$, and $\lambda_{\rm{AA}}$ are hyper parameters for balancing the losses. The effects of each hyper parameter are detailed in the supplementary materials.

\section{Experiments}\label{sec5}
\subsection{Experimental setting}\label{sec5-1}
\textbf{Dataset.}
To facilitate IID using LiDAR intensity during training, we employ the NTT-IID dataset~\cite{Sato2023}, which consists of images, LiDAR intensities, and annotations designed for evaluating IID quality. The NTT-IID dataset prepares 10,000 pairs of images measuring outdoor scenes and LiDAR intensity mapped to these images. Among these pairs, 110 samples have been annotated, yielding a total of 12,626 human judgments. Additionally, we utilized the FSVG dataset~\cite{krahenbuhl2018} as the target domain for albedo and shade. In this paper, we employ the same albedo and shade samples as those used in IID-LI~\cite{Sato2023}.

\newpage

\tabcolsep = 2pt
\begin{table}
\centering
\scalebox{0.78}{
\begin{tabular}{ccccccccccc}
\toprule
\multirow{2}{*}{Model} & \multicolumn{4}{c}{Random sampled annotation}& \multicolumn{1}{c}{}& \multicolumn{4}{c}{All annotation}\\
\cmidrule(lr){2-5}\cmidrule(lr){7-10}
&F-score($\uparrow$)&WHDR($\downarrow$)&Precision($\uparrow$)&Recall($\uparrow$)&
&F-score($\uparrow$)&WHDR($\downarrow$)&Precision($\uparrow$)&Recall($\uparrow$)\\
\midrule
Baseline-R~\cite{bell2014}&0.350&0.527&0.375&0.440&&0.306&0.531&0.393&0.445\\
Baseline-S~\cite{bell2014}&0.227&0.529&0.361&0.340&&0.314&\color{blue}0.185&0.431&0.340\\
Retinex~\cite{grosse2009}       &0.420&0.452&0.523&0.445&&0.469&0.187&0.496&0.455\\
Color Retinex~\cite{grosse2009} &0.420&0.452&0.531&0.445&&0.470&0.187&0.496&0.455\\
Bell et al.~\cite{bell2014}     &0.414&0.446&0.504&0.453&&0.457&0.213&0.467&0.463\\
Bi et al.~\cite{bi2015}         &0.490&0.406&0.561&0.522&&0.466&0.283&0.462&0.522\\
Revisiting$^*$~\cite{fan2018}&0.442&0.428&\color{blue}0.635&0.470&&0.499&\color{red}0.181&\color{red}0.575&0.485\\
IIDWW~\cite{li2018}             &0.417&0.464&0.489&0.475&&0.397&0.375&0.418&0.483\\
UidSequence~\cite{Lettry2018}   &0.419&0.483&0.453&0.450&&0.395&0.372&0.405&0.453\\
USI$^3$D~\cite{liu2020}         &0.454&0.422&0.539&0.500&&0.446&0.287&0.444&0.504\\
IID-LI~\cite{Sato2023}          &\color{blue}0.602&\color{blue}0.353&0.625&\color{blue}0.596&
&\color{blue}0.521&0.227&\color{blue}0.517&\color{blue}0.591\\
LIET (Ours)            &\color{red}0.607&\color{red}0.340&\color{red}0.649&\color{red}0.601&
&\color{red}0.525&0.245&0.500&\color{red}0.598\\
\bottomrule
\end{tabular}
}
\caption{\label{table1} Numerical comparison in IID quality with NTT-IID dataset~\cite{Sato2023}. Due to the bias in the number of annotations, we evaluated (i) randomly sampled annotation and (ii) all annotations along with IID-LI~\cite{Sato2023}. Red and blue fonts indicate the best and second-best results, respectively. For both randomly sampled annotation and all annotations, LIET (Ours) achieves comparable IID quality to IID-LI. In addition, Revisiting$^*$~\cite{fan2018} demonstrates the better IID quality in two indices of all annotations. Due to biases in the distribution of all annotations, models that infers flatter albedos are at an advantage in these metrics. Revisiting$^*$~\cite{fan2018}, assuming local flatness of albedo in its training, demonstrates superior results by aligning with this bias.}
\end{table}

\vspace{-0.5cm}
\noindent
\textbf{Evaluation metric.}
For quantitative evaluation, we employ the following metrics: the weighted human disagreement rate (WHDR), precision, recall, and F-score for all and random sampled annotation, following the same methodology as IID-LI~\cite{Sato2023}. In addition, we evaluate the image quality using five IQA models: MANIQA~\cite{NIQA}, TReS~\cite{TReS}, MUSIQ~\cite{MUSI}, HyperIQA~\cite{HIQA}, and DBCNN~\cite{DBCN}.\smallskip\\
\textbf{Implementation details.}
In LIET, the style code encoder $E^p_{\rm{X}}$, content code encoder $E^c_{\rm{X}}$, generator $G_{\rm{X}}$, and discriminator $D_{\rm{X}}$, $(X\in{I, L, R, S})$, are implemented with the same model structures and parameters as USI$^3$D~\cite{liu2020}\footnote{LIET is implemented based on USI$^3$D
(\href{https://github.com/DreamtaleCore/USI3D.git}{https://github.com/DreamtaleCore/USI3D.git})} and IID-LI. The style encoders $E^p_{\rm{X}}$, the content code encoders $E^c_{\rm{X}}$, the decoders $G_{\rm{X}}$, and the discriminators $D^c_{\rm{X}}$ consist of convolutional layers, down-sampling layers, global average pooling layers, dense layers, and residual blocks. Notably, the residual blocks incorporate adaptive instance normalization (AdaIN)~\cite{huang2017}, and the AdaIN parameters are dynamically determined using a multi-layer perceptron (MLP). To assess images from both global and local perspectives, a multi-scale discriminator~\cite{wang2018high} is employed in discriminator $D_{\rm{X}}$. Given a style code, style mapping modules $M_{\rm{I}}$ and $M_{\rm{L}}$ are constructed by MLP. We empirically set the following values for the hyper parameters: $\lambda_{\rm{img}} = 100.0, \lambda_{sty} = 10.0, \lambda_{cnt} = 1.0, \lambda_{KLD} = 1.0, \lambda_{VGG} = 1.0, \lambda_{phy} = 10.0$, and $\lambda_{AA} = 100.0$.

\subsection{Intrinsic image decomposition quality}\label{sec5-2}
LIET is compared with energy optimization models including baseline-R~\cite{bell2014}, baseline-S~\cite{bell2014}, Retinex~\cite{grosse2009}, Color Retinex~\cite{grosse2009}, Bell et al.~\cite{bell2014}, and Bi et al.~\cite{bi2015}. 

\newpage

\begin{figure}
\centering
\includegraphics[width=1.0\linewidth]{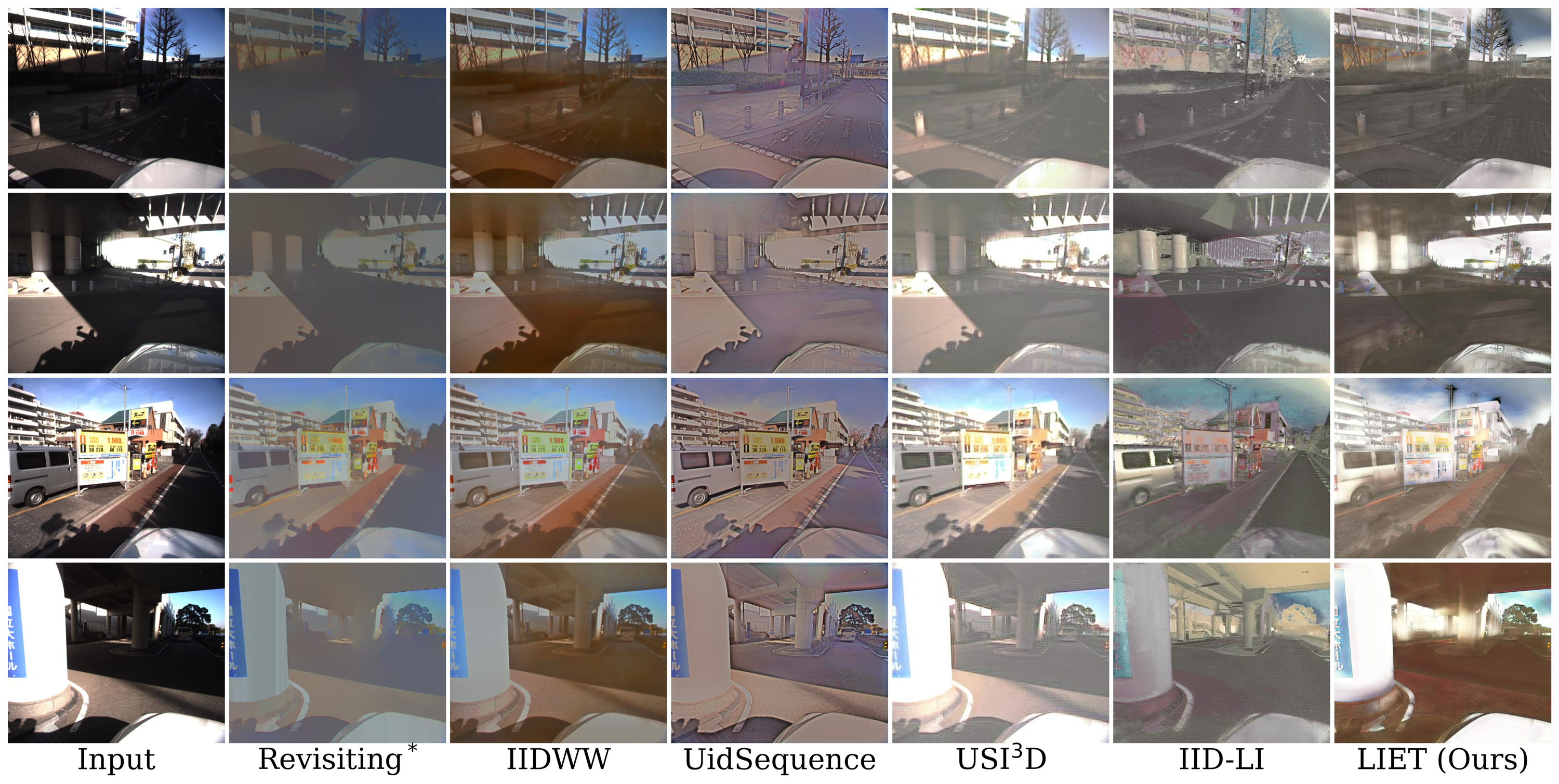}
\caption{\label{fig6} Examples of inferring results obtained from various existing models and LIET (Ours) with NTT-IID dataset~\cite{Sato2023}. The compared models include Revisiting$^*$~\cite{fan2018}, IIDWW~\cite{li2018}, UidSequence~\cite{Lettry2018}, USI$^3$D~\cite{liu2020}, and IID-LI~\cite{Sato2023}. Shadows are less noticeable on the IID-LI and LIET, while cast shadows are visibly retained on the existing models without LiDAR intensity utilization.}
\end{figure}

\vspace{-0.5cm}
\noindent
Additionally, Revisiting$^*$\footnote{Revisiting~\cite{fan2018} is marked with an asterisk ($^*$) since it is a supervised learning model trained with the IIW dataset~\cite{bell2014}.The IIW dataset features real-world images with human-judged annotations. Unlike the NTT-IID dataset~\cite{Sato2023}, it lacks LiDAR intensity but stands out for its large quantity of data and annotations.}~\cite{fan2018} is also evaluated as a supervised model. 
Furthermore, IIDWW~\cite{li2018}, UidSequence~\cite{Lettry2018}, USI$^3$D~\cite{liu2020}, and IID-LI~\cite{Sato2023} are implemented as unsupervised learning models. \cref{table1} shows the numerical results for the NTT-IID dataset. 
Consequently, LIET demonstrates a comparable performance to  that of IID-LI which utilizes a single image and its corresponding LiDAR intensity during inference, despite inputting only a single image in LIET. With all annotation, Revisiting$^*$~\cite{fan2018} performed better on the two metrics. This model incorporates an albedo flattening module, and these metrics are more favorable for inferring flat albedos due to the annotation bias. Subsequently, the qualitative results of existing models and LIET are illustrated in~\cref{fig6}. In comparison to other IID models, both IID-LI and LIET yield inferred albedos with less noticeable shadows due to the LiDAR intensity utilization. Though Revisiting~\cite{fan2018} exhibits a flattened appearance, leading to favorable quantitative outcomes, cast shadows within the images still remain.

\subsection{Image quality}\label{sec5-3}
Subsequently, we conducted an IQA comparison between LIET and the top three models for their IID quality: Revisiting$^*$~\cite{fan2018}, USI$^3$D~\cite{liu2020}, and IID-LI~\cite{Sato2023}. As shown in~\cref{table2}, LIET achieves the highest quality across all five evaluation metrics. Revisiting$^*$~\cite{fan2018} is trained with relative gray-scaled albedo between nearby

\newpage

\tabcolsep = 2pt
\begin{table}
\centering
\scalebox{0.78}{
\begin{tabular}{@{}cccccc@{}}
\toprule
Model &MANIQA($\uparrow$)&TReS($\uparrow$)&MUSIQ($\uparrow$)&HYPERIQA($\uparrow$)&DBCNN($\uparrow$)\\
\midrule
Input  &(0.664)&(81.1)&(59.2)&(0.595)&(58.1)\\
Revisiting$^*$~\cite{fan2018} &0.395&55.4&\color{blue}44.9&\color{blue}0.389&\color{blue}38.1\\
USI$^3$D~\cite{liu2020}&\color{blue}0.488&53.1&40.1&0.298&35.3\\
IID-LI~\cite{Sato2023}       &0.460&\color{blue}55.5&43.5&0.285&37.5\\
LIET (Ours)         &\textcolor{red}{0.570}&\textcolor{red}{75.1}&\color{red}56.3&\color{red}0.414&\color{red}46.3\\
\bottomrule
\end{tabular}
}
\caption{\label{table2} Numerical comparison in IQA between LIET and the top three models in IID quality including Revisiting$^*$~\cite{fan2018}, USI$^3$D~\cite{liu2020}, and IID-LI~\cite{Sato2023}. Our findings demonstrate that the images inferred by LIET consistently exhibited the highest image quality across all metrics. This comparable performance can be attributed to the absence of image blurring and collapse.}
\end{table}

\vspace{-1.0cm}

\begin{table}
\centering
\scalebox{0.78}{
\begin{tabular}{ccccccccccc}
\toprule
\multirow{2}{*}{Model} & \multirow{2}{*}{$\mathcal{L}^{\rm{smooth}}$} & \multicolumn{4}{c}{IID quality metrics}& \multicolumn{1}{c}{}& \multicolumn{3}{c}{IQA metrics}\\
\cmidrule(lr){3-6}\cmidrule(lr){8-10}
&&F-score($\uparrow$)&WHDR($\downarrow$)&Precision($\uparrow$)&Recall($\uparrow$)&&MANIQA($\uparrow$)&TReS($\uparrow$)&MUSIQ($\uparrow$)\\
\midrule
USI$^3$D~\cite{liu2020}       &               &0.444&0.427&0.540&0.502&&0.629&67.2&52.6\\
USI$^3$D$^{**}$~\cite{liu2020} &$\checkmark$   &0.454&0.422&0.539&0.500&&0.492&53.2&40.1\\
IID-LI~\cite{Sato2023}        &               &0.570&0.389&0.581&0.565&&0.522&80.9&56.4\\
IID-LI$^{**}$~\cite{Sato2023}  &$\checkmark$   &0.602&0.353&0.625&0.596&&0.461&55.5&43.5\\
LIET$^{**}$ (Ours)             &               &0.607&0.340&0.649&0.601&&0.570&75.1&56.3\\
LIET (Ours)                   &$\checkmark$   &0.593&0.368&0.634&0.583&&0.556&69.5&50.1\\
\bottomrule
\end{tabular}
}
\caption{\label{table3} Effect of smoothing loss $\mathcal{L}^{\rm{smooth}}$ in IID quality and IQA with the NTT-IID dataset~\cite{Sato2023}. Smoothing loss improves the IID quality for USI$^3$D and IID-LI, while degrading image quality due to its blurring effect. LIET has already achieved albedo local flatness, hence adding smoothing loss will not improve the IID quality.}
\end{table}

\vspace{-0.5cm}
\noindent
points. Thus the model output tends to reduce saturation and leads to reducing IQA ratings. Additionally, both USI$^3$D and IID-LI enhance IID quality through a smoothing process that assumes local albedo flatness. As a result, the images inferred by these models tend to exhibit blurriness, leading to lower IQA ratings. Conversely, in LIET, since fine shadows derived from cast shadows are absent in LiDAR intensity, the albedo inferred from LiDAR intensity has less variation in luminance. LIET achieves albedo local flatness without using smoothing loss due to the alignment of the albedo inferred from LiDAR intensity with the albedo inferred from the image by albedo-alignment loss. The effect of smooth loss is described in the next section.

\subsection{Effect of smoothing loss}
Smoothing loss is not implemented in LIET, despite USI$^3$D~\cite{liu2020} and IID-LI~\cite{Sato2023} utilizing it to improve IID quality. Thus, this section describes the effect of smoothing loss for IID quality and IQA, by removing and adding the smooth loss for USI$^3$D~\cite{liu2020}, IID-LI~\cite{Sato2023}, and LIET~\footnote{USI$^3$D with smooth loss, IID-LI with smooth loss, and LIET without smooth loss are marked with a double asterisk ($^{**}$) since these models are original}. As a result, for both USI$^3$D and IID-LI, eliminating smoothing loss improves the image quality of the inferred albedos as shown in~\cref{table3}, since the blurring by smoothing loss is reduced. On the other hand, variation caused by cast shadows in the albedo was reduced by 

\newpage

\begin{table}
\centering
\scalebox{0.78}{
\begin{tabular}{ccccc}
\toprule
Model&F-score($\uparrow$)&WHDR($\downarrow$)&Precision($\uparrow$)&Recall($\uparrow$)\\
\midrule
Ours &0.607&0.340&0.649&0.601\\
w/o $\mathcal{L}^{\rm{AA}}$ &0.437&0.473&0.497&0.476\\
w/o inst.&0.489&0.447&0.520&0.505\\
w/o gray  &0.601&0.359&0.623&0.596\\
w/o ILC path&0.589&0.361&0.641&0.581\\
\bottomrule
\end{tabular}
}
\caption{\label{table4}Effect of albedo-alignment loss $\mathcal{L}^{\rm{AA}}$. "w/o inst." describes the loss $\mathcal{L}^{\rm{AA}}$ calculated without instance normalization in albedo-alignment loss. "w/o gray" refers to delete the gray scaling from albedo-alignment loss.}
\end{table}

\vspace{-0.5cm}
\noindent
smoothing, hence, IID quality degrades without smoothing loss. Conversely, the albedo inferred from LiDAR intensity has less variability derived from cast shadows, hence the variation in the albedo inferred from the image has already reduced due to albedo-alignment loss in LIET. Thus, LIET achieves albedo local flatness without using smoothing loss, which keeps the IQA closely aligned with the input image. Adding further smoothing only degrades IQA and does not improve IID quality.

\subsection{Ablation study}\label{sec5-4}
This section describes an ablation study for the contribution of albedo-alignment loss and ILC paths. As shown in~\cref{table4}, the albedo-alignment loss is effective for LIET due to the direct connection between the image-encoder path and the LiDAR-encoder path during training. Since the distribution of LiDAR intensity varies across samples, features are well-trained by applying instance normalization rather than by scaling uniformly across all samples. The inference quality is slightly improved by aligning these albedos in gray scale, due to the lack of hue in LiDAR intensity. Additionally, the ILC paths contributes the separating an image into content and style codes by mutually translating the image and its corresponding LiDAR intensity, which share the contents but differ styles. Thus, the IID quality is improved by ILC paths.

\section{Conclusion}\label{sec6}
In this paper, we proposed \textit{unsupervised single-image intrinsic image decomposition with LiDAR intensity enhanced training (LIET)}. We proposed a novel approach in which an image and its corresponding LiDAR intensity are individually fed into the model during training, while the inference process only employs a single image. To calculate the relationship between the image-encoder path and the LiDAR-encoder path, we introduced albedo-alignment loss to align the albedo inferred from a single image to that from its corresponding LiDAR intensity, and ILC paths to enhance the separation of contents and styles. As a result, LIET achieved performance comparable to state-of-the-art in IID quality metrics while only employing a single image as input during inference. Furthermore, LIET demonstrated improvements in image quality supported by the five most recent IQA metrics.

%
%
\bibliographystyle{splncs04}
\bibliography{main}
\end{document}